\pdfoutput=1

\documentclass[11pt]{article}

\usepackage[final]{EMNLP2023}

\usepackage{times}
\usepackage{latexsym}

\usepackage[T1]{fontenc}

\usepackage[utf8]{inputenc}

\usepackage{microtype}

\usepackage{inconsolata}
\usepackage{hyperref}       
\usepackage{url}            
\usepackage{booktabs}       
\usepackage{amsfonts}       
\usepackage{nicefrac}       
\usepackage{xcolor}         
\usepackage{graphicx}
\usepackage{caption}
\usepackage{subcaption}
\usepackage{wrapfig}
\usepackage{colortbl}
\usepackage{epsfig}
\usepackage{graphicx}
\usepackage{float}
\usepackage{wrapfig}
\usepackage{algorithm,algorithmicx,algpseudocode}
\usepackage{bm,xspace}
\usepackage{comment}
\usepackage{verbatim}
\usepackage{multirow}
\usepackage{array}
\usepackage{balance}
\usepackage{url}
\usepackage{booktabs}
\usepackage{etoolbox,siunitx}
\usepackage{calc}
\usepackage{pifont,hologo}
 
\usepackage{nicefrac}

\usepackage{arydshln}
\usepackage[utf8]{inputenc}
\usepackage[T1]{fontenc}
\usepackage{url} 
\usepackage{amsfonts}
\usepackage{nicefrac}
\usepackage{microtype}
\usepackage[american]{babel}
\usepackage{enumitem}
\usepackage{amssymb}
\usepackage{amsmath}
\usepackage{siunitx}

\usepackage[super]{nth}
\usepackage{nicefrac}
\robustify\bfseries

\usepackage{dsfont}

\DeclareMathSymbol{@}{\mathord}{letters}{"3B}

\makeatletter
\DeclareRobustCommand\onedot{\futurelet\@let@token\@onedot}
\def\@onedot{\ifx\@let@token.\else.\null\fi\xspace}

\def\ie{\emph{i.e}\onedot} 
 
 \def\vs{\emph{vs}\onedot}
\def\wrt{w.r.t\onedot} 

\makeatother

\newcommand{\vlnlong}{Vision-and-Language Navigation\xspace}
\newcommand{\vln}{VLN\xspace}

\newcommand{\rtor}{Room-to-Room\xspace}
\newcommand{\rxr}{Room-across-Room\xspace}

\newcommand{\shuffling}{Nonsensical Instructions\xspace}

\newcommand{\sfwordlong}{Word order within sentence\xspace}
\newcommand{\sfsentlong}{Sentence order\xspace}
\newcommand{\sfwordsentlong}{Word order within sentence + sentence order\xspace}
\newcommand{\sfallwordlong}{All words shuffled\xspace}

\newcommand{\orig}{\textsc{Original}\xspace}
\newcommand{\sfword}{\textsc{SF-word}\xspace}
\newcommand{\sfsent}{\textsc{SF-sent}\xspace}
\newcommand{\sfwordsent}{\textsc{SF-word-sent}\xspace}
\newcommand{\sfallword}{\textsc{SF-all}\xspace}

\newcommand{\permutation}{Irrelevant Instructions\xspace}

\newcommand{\permblocklong}{mismatch as blocks\xspace}
\newcommand{\permalllong}{mismatch randomly\xspace}

\newcommand{\permblock}{\textsc{Mismatch-B}\xspace}
\newcommand{\permall}{\textsc{Mismatch-R}\xspace}

\newcommand{\ugtitle}{Unigram + Object\xspace}
\newcommand{\uglong}{Unigram + Object\xspace}
\newcommand{\ug}{\textsc{UO}\xspace}

\newcommand{\trainonly}{\textsc{R2R-only}\xspace}
\newcommand{\augonly}{\textsc{Aug-only}\xspace}
\newcommand{\augonlytrainsize}{\textsc{Aug-R2R-size}\xspace}
\newcommand{\trainandaug}{\textsc{R2R\&Aug}\xspace}
\newcommand{\rxraug}{\textsc{Marky}\xspace}

\newcommand{\vilb}{ViLBERT\xspace}
\newcommand{\hamt}{HAMT\xspace}
\newcommand{\vlnb}{VLN-BERT\xspace}

\newcommand{\airb}{Airbert\xspace}

\newcommand{\mlmlong}{masked language modeling\xspace}
\newcommand{\mrmlong}{masked region modeling\xspace}
\newcommand{\itmlong}{instruction
trajectory matching\xspace}
\newcommand{\saprlong}{single-step action prediction/regression\xspace}
\newcommand{\sprellong}{spatial relationship prediction\xspace}

\newcommand{\mlm}{MLM\xspace}
\newcommand{\mrm}{MRM\xspace}
\newcommand{\itm}{ITM\xspace}
\newcommand{\sap}{SAP\xspace}
\newcommand{\sapr}{SAP/SAR\xspace}
\newcommand{\sprel}{SPREL\xspace}

\newcommand{\sr}{\textsc{SR}\xspace}
\newcommand{\spl}{\textsc{SPL}\xspace}
\newcommand{\asr}{\textsc{SRA}\xspace}
\newcommand{\ndtw}{n\textsc{DTW}\xspace}

\newcommand{\agreementlong}{success rate agreement\xspace}
\newcommand{\agreement}{\textsc{SRA}\xspace}

\newcommand{\dc}{Data Cartography\xspace}

\title{Does VLN Pretraining Work with Nonsensical or Irrelevant Instructions?}

\author{Wang Zhu \quad\quad Ishika Singh\thanks{\xspace\xspace Author contributed equally} \quad\quad Yuan Huang{\footnotemark[1]} \quad\quad Robin Jia \quad\quad Jesse Thomason \\ \\
University of Southern California, Los Angeles, CA, USA \\
\texttt{\{wangzhu, ishikasi, yuanhuan, robinjia, jessetho\}@usc.edu}}

\begin{document}
\maketitle
\begin{abstract}
Data augmentation via back-translation is common when pretraining \vlnlong (\vln) models, even though the generated instructions are noisy.
But: does that noise matter?
We find that nonsensical or irrelevant language instructions during pretraining can have little effect on downstream performance for both \hamt and \vlnb on R2R, and is still better than only using clean, human data.
To underscore these results, we concoct an efficient augmentation method, \uglong, which generates nonsensical instructions that nonetheless improve downstream performance.
Our findings suggest that what matters for \vln R2R pretraining is the \emph{quantity} of visual trajectories, not the \emph{quality} of instructions.
\end{abstract}

\begin{figure}
    \centering
    \includegraphics[width=\linewidth]{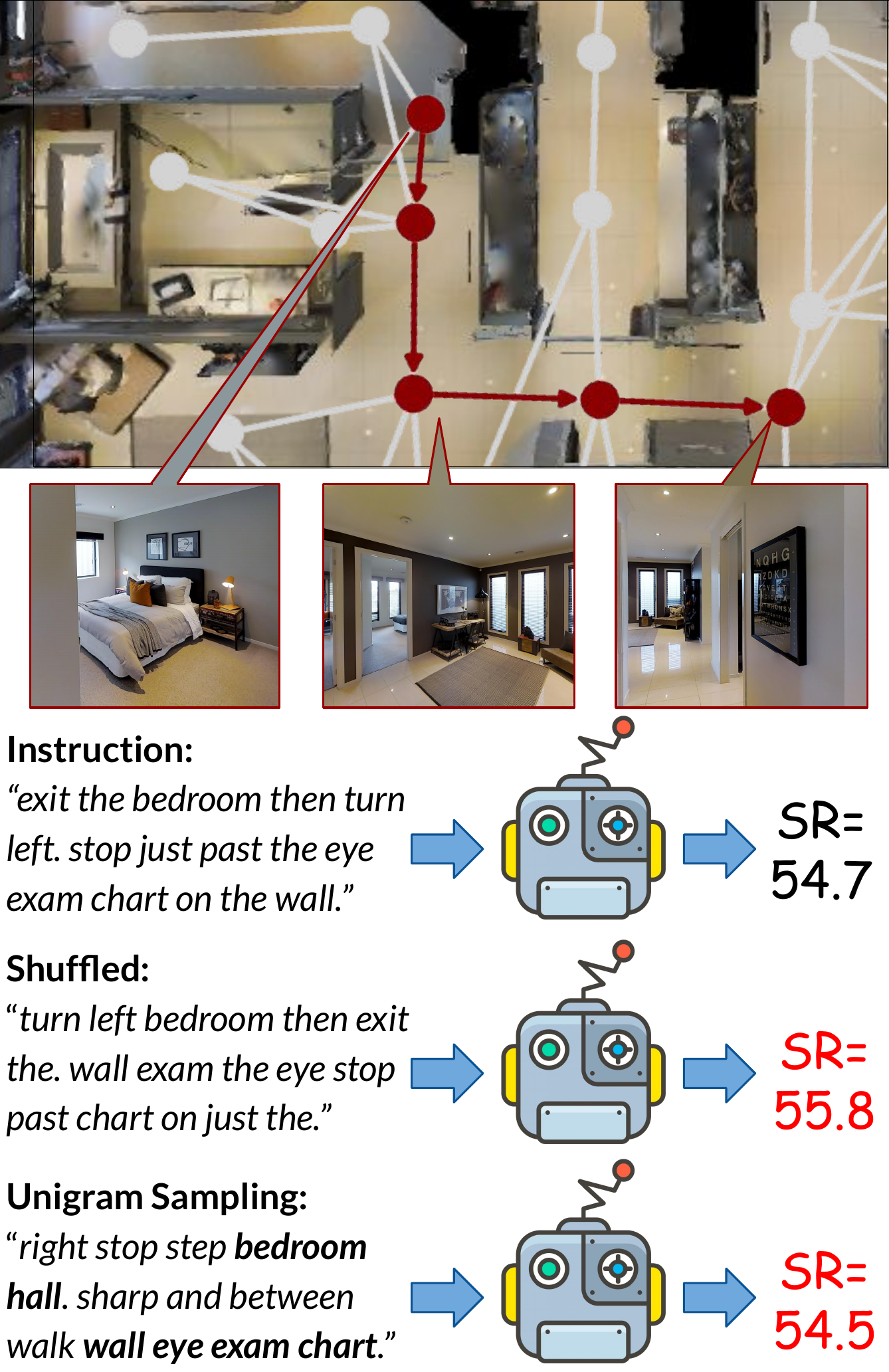}
    \caption{We find that pretraining with trajectories paired with instructions with shuffled words, or cheaply generated instructions consisting of object words with meaningless filler can match downstream performance compared to pretraining with human instructions.}
    \label{fig:overview}
\end{figure}


\section{Introduction}
\label{sec:intro}
\vlnlong is a multimodal instruction-following task.
An agent navigates to a goal in a simulated environment based on a natural language instruction and visual observations.
Existing \vln datasets like R2R~\cite{anderson2018vision} have wide visual diversity across environments and objects, but only a small number of human-annotated language instructions covering a small fraction of the possible trajectories in the environment.
To compensate, prior work on pretraining for \vln augments the training data with synthetic instructions generated by a speaker model~\cite{fried2018speaker} given an image trajectory. 
These generated instructions are known to be noisy and often unrelated to the path~\cite{huang2019transferable}, but they are nonetheless helpful for \vln pretraining.

In light of these observations, what quality of data is sufficient for \vln pretraining? Could even nonsensical or irrelevant data be useful?

To understand why the noisy augmentation is sufficient for \vln pretraining, we first test two extreme noising strategies to create nonsensical or irrelevant instructions for pretraining data.
Across two strong transformer-based models, \hamt and \vlnb, shuffling words or sentences has little negative impact on final navigation success in Room-2-Room (R2R)~\cite{anderson2018vision}.
Pretraining on randomly paired trajectories and instructions has more of a negative effect, but still greatly outperforms not pretraining at all.
Based on these observations, we devise a simple and efficient method, \uglong (\ug), to generate nonsensical augmentation data that is still useful for \vln pretraining, without a seq-to-seq speaker model.
\ug generates an ``instruction'' for a trajectory by concatenating names of detected objects along the trajectory with random \vln-related words. 
For \hamt, pretraining on a mixture of \ug nonsensical data and irrelevant instructions outperforms pretraining on the original R2R data; for \vlnb, pretraining on \ug nonsensical data alone is on par with pretraining on human-annotated training data.

We also experiment with changing the \emph{quantity} of pretraining data, by using either only the high-quality training data or only the noisy augmentation data.
Pretraining with only the augmentation data performs better than using only the training data, and is nearly on par with using both data sources. 
We conclude that data quantity matters more than quality in \vln pretraining. 
In particular, since the augmentation data covers many more visual trajectories, it helps the model learn image-based pretraining tasks even though the instructions are noisy.
Overall, our findings suggest that what matters most for \vln R2R pretraining is having a large number of trajectories, even if they are paired with nonsensical or irrelevant language instructions.

\begin{figure*}[t]
    \centering
    \includegraphics[width=0.95\textwidth]{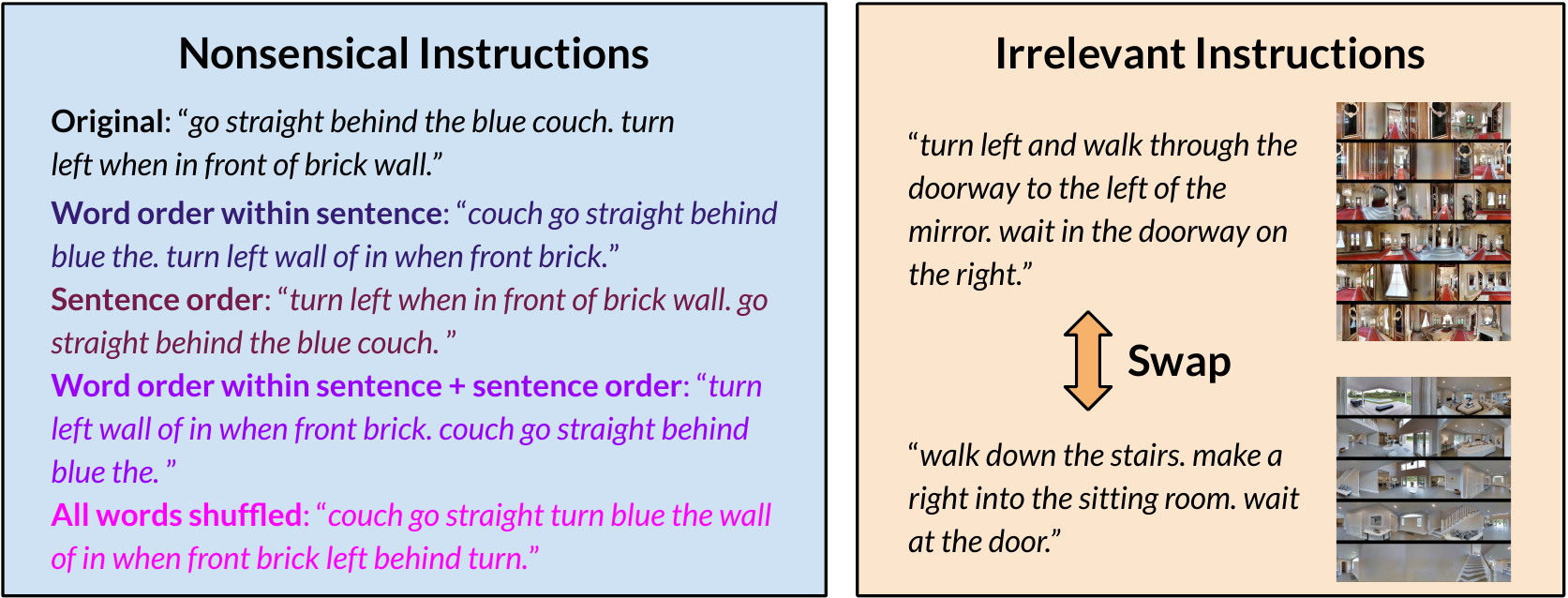}
    \caption{We explore two noising strategies: shuffling word orders within each annotation instruction (left) and pairing language annotations with trajectories to which they do not correspond (right).
    Surprisingly, we find that pretraining data having undergone either noising strategy can support decent downstream performance.
    }
    \label{fig:nosing}
\end{figure*}

\section{Related Works}
\label{sec:related}

\paragraph{\vln pretraining} 
\vlnb~\cite{majumdar2020improving} first applied large transformer-based model for \vln pretraining, yielding significant gains over prior models in downstream R2R success.
\citet{hao2020prevalent} improved over \vlnb using self-supervised learning on image-text-action triplets.
Further,~\citet{chen2021hamt} proposed an advanced transformer-based model \hamt, that encodes the navigation history additionally and designed proxy tasks to pretrain it. 
\hamt improves \vln performance on several downstream tasks, including R2R~\cite{mattersim}, REVERIE~\cite{reverie}, and CVDN~\cite{thomason:corl19}.
\airb~\cite{guhur2021airbert} introduced a new pretraining dataset for \vln built over images and descriptions of indoor houses.
Because the pretraining data in \airb is unavailable, our analysis focuses on \vlnb and \hamt.

\paragraph{Data augmentation for \vln}
~\citet{fried2018speaker} developed a method that augments the paired instruction and demonstration data using a learned speaker model.
Such augmentation instructions can have little connection to the trajectories they are meant to describe~\cite{huang2019transferable}.
To improve the quality of augmentation,~\citet{wang2021more} trained a landmark detector and applied a multilingual T5-based language generator to generate high-quality language instructions.
On the other hand, ~\citet{tan2019learning} and \citet{li2022envedit} applied the ``environmental dropout'' for visual augmentation through masking or editing specific types of objects during training.
These visual augmentation methods are widely used in the current \vln pretraining pipelines. 
Unlike them, we investigate methods to achieve effective pretraining performance via even noisier augmentation that requires substantially less engineering and bespoke model assembly.

\paragraph{Pretraining data quality in NLP}
~\citet{shuffle} found that word order does not matter for a number of downstream language-only tasks. 
Relatedly, ~\citet{krishna-etal-2021-pretraining-summarization, papadimitriou-jurafsky-2020-learning} finds that nonsense text or unrelated out-of-distribution data in pretraining help with success in downstream tasks. 
These findings inspire us to explore the sensitivity on word order of \vln, and design the noisier augmentation method for \vln pretraining.

\paragraph{\vln analysis} 
Closest in spirit to our work, \citet{zhu-etal-2022-diagnosing} diagnose the decision-making in \vln agents by perturbing the inference-time input to fully trained VLN agents.
They mask out parts of the instruction and the visual observation, such as object tokens, bounding boxes, and direction tokens, to understand the kind of information that the model attends to \emph{during evaluation}.
By contrast, in this work we study the effects of \textit{pretraining} and data augmentation for \vln agents under different settings, rather than testing inference behavior of off-the-shelf, fully-trained agents.


\section{Task and Setups}
\label{sec:task}

In \vln, the agent receives a natural language instruction describing a path towards a goal location and predicts a sequence of actions reach that goal.
We overview the dataset, models, and evaluation measures we use to analyze the effects of noising instructions during pretraining and fine-tuning.

\subsection{Dataset}
\label{subsec:dataset}
We focus on the \rtor (R2R; \citealp{anderson2018vision}) benchmark dataset of human instructions and corresponding trajectories, as well as augmentation data of trajectories with machine-generated instructions. 
R2R has 4.6k trajectories and 14k human instructions as training data in 61 indoor scenes, and 2.3k trajectories and 7k human instructions as the validation data in 11 scenes unseen at training time.
Each trajectory has 3 or 4 language instruction annotations.
R2R trajectories are defined in a simulated topological graph environment, and all the trajectories finish in 3 to 6 steps through the graph.
We additionally consider an augmented dataset~\cite{fried2018speaker} that provides model-based predictions of language annotations for all possible 170k trajectories of length 3 to 6 in the 61 training scenes.
We use the largest publicly available speaker augmentation from~\citet{hao2020prevalent}, which has 6 instruction annotations per unannotated trajectory, resulting in around 1 million instruction-trajectory pairs.

\begin{table*}[t]
\setlength{\aboverulesep}{0pt}
\setlength{\belowrulesep}{0pt}
\centering
\begin{tabular}{l>{\columncolor[gray]{0.8}}lrrrrr}
\vln Model & Pretraining Shuffle & \multicolumn{5}{c}{\cellcolor{blue!20}Finetuning Shuffle} \\
& & \cellcolor{blue!20}{\small\orig} & \cellcolor{blue!20}{\small\sfword} & \cellcolor{blue!20}{\small\sfsent} & \cellcolor{blue!20}{\small\sfwordsent} & \cellcolor{blue!20}{\small\sfallword} \\
\toprule
\multirow{5}{*}{\textbf{\hamt}} 
& {\small\orig} & $57.8$ & $56.0$ & $54.6$ & $52.3$ & $53.1$ \\  
& {\small\sfword} & $58.1$ & $56.6$ & $54.2$ & $53.8$ & $55.4$ \\  
& {\small\sfsent} & $56.3$ & $54.2$ & $55.7$ & $53.3$ & $52.1$ \\  
& {\small\sfwordsent} & $54.9$ & $51.2$ & $50.5$ & $46.8$ & $48.0$ \\  
& {\small\sfallword} & $55.9$ & $52.6$ & $51.0$ & $49.8$ & $50.0$ \\  
\midrule
\multirow{5}{*}{\textbf{\vlnb}} 
& {\small\orig} & $54.7$ & $53.5$ & $52.1$ & $45.6$ & $47.6$ \\
& {\small\sfword} & $55.8$ & $55.9$ & $52.6$ & $47.0$ & $51.0$ \\
& {\small\sfsent} & $54.7$ & $52.4$ & $53.3$ & $49.5$ & $51.0$ \\
& {\small\sfwordsent} & $53.0$ & $53.3$ & $52.8$ & $50.6$ & $48.3$ \\
& {\small\sfallword} & $54.0$ & $54.8$ & $51.6$ & $48.5$ & $48.3$\\
\bottomrule
\end{tabular}
\caption{
The \sr on R2R validation unseen set when shuffling training instructions.
\hamt is finetuned with IL and averaged over 3 runs.
The results show (1) increased noising decreases \sr [rows], and (2) only severe noising in pretraining starts having effects in final performance [columns], especially when both pretrain \& finetune are noised.
}
\label{tab:shuffling}
\end{table*}

\subsection{Models and Pretraining Tasks}
\label{subsec:exp_setup}
We conduct experiments with \hamt~\cite{chen2021hamt} and \vlnb~\cite{majumdar2020improving}.

\textbf{\hamt}~\cite{chen2021hamt} is a transformer-based agent trained to predict next navigation action in a closed-loop fashion.
\hamt consists of three transformer encoders: a language encoder, a visual encoder for the current observation, and a history encoder for previous state-action pairs; and finally a cross-modal transformer decoder that fuses these to predict the next action.
The model is randomly initialized, then pretrained using the R2R training split and the 1 million noisy speaker-augmentated instructions from \citet{hao2020prevalent} and fine-tuned with only the R2R training data.
In addition to used proxy tasks in vision-language pretraining, i.e, \mlmlong (\mlm), \mrmlong (\mrm) and \itmlong (\itm), \hamt adds another three tasks during its pretraining phase.
Two pretraining tasks, \saprlong (\sapr), are grounding-based, as they require the model to predict the next action based on the instruction.
The other pretraining task, \sprellong (\sprel), only involves visual input; the model must predict the relative spatial position of two views.
Following~\citet{chen2021hamt}, during fine-tuning we train using imitation learning (IL) with teacher-forcing and reinforcement learning (RL) with student-forcing.

\textbf{\vlnb}~\cite{majumdar2020improving} is trained to score candidate paths to select the final set of actions.
Candidate paths are sampled from beam search by a LSTM-based follower agent~\cite{fried2018speaker}.
\vlnb is initialized with \vilb~\cite{lu2019vilbert}, 
then further pretrained with \mlm and \mrm as proxy tasks.
R2R training data is used for pretraining.
\vlnb is fine-tuned on the path scoring task.
We apply the same pretraining and fine-tuning setup for \vlnb.

\subsection{Evaluation Measures}
We report success rate (\sr) and success weighted by path length (\spl), the main ranking metrics on the R2R leaderboard.
Additionally, we report the normalized dynamic time warping (\ndtw)~\cite{magalhaes2019effective} metric in Appendix~\ref{secapp:exps}.
We report the results on the validation unseen scenes.

To measure the similarity of two models, we define the \agreementlong (\agreement). 
For each instruction in the validation unseen set, we set the agreement per instruction as 1 if the SR of the predictions from two models are the same, and 0 otherwise. 
We then average agreement per instruction over all the instructions to compute the \agreement. Suppose there are $n$ instructions in the validation unseen set, the \agreement between models $X$ and $Y$ is
\begin{align*}
    \asr(X,Y) = \frac{1}{n}\sum_i\mathbb{I}[\sr^X_i = \sr^Y_i]
\end{align*}
for $\sr^X_i$ the \sr on instruction $i$ for model $X$.


\section{Noising Strategies}
\label{sec:strategy_r2r}

We discover that high-quality language annotations are not necessary to achieve performance gains during \vln pretraining over using human-annotated data alone.
We introduce several noising strategies over the existing instructions to create nonsensical or irrelevant instructions.

\subsection{\shuffling}
\label{ssec:shuffling}

We apply different word shuffling strategies within each instruction that preserve the mapping between trajectories and instructions during pretraining and fine-tuning but introduce word-order noise in the instructions themselves (Fig.~\ref{fig:nosing}, left).
Note that instructions are tokenized by the spaCy Python library and words are not split into multiple tokens.

(1)~\sfwordlong (\sfword):
Each instruction contains 1 to 3 sentences separated by periods. 
We shuffle words within each sentence but preserve the sentence order.

(2)~\sfsentlong (\sfsent):
We preserve word order within each sentence but shuffle the order of sentences in the instructions, not allowing the original order.
Only instructions with only one sentence are unaffected; over 85\% of the training data is changed via sentence re-ordering.

(3)~\sfwordsentlong (\sfwordsent):
We shuffle \textit{both} the word order within sentences and the sentence order itself.

(4)~\sfallwordlong (\sfallword):
We remove punctuation tokens and shuffle the remainder, appending a single period to the resulting instruction.
 
We apply these nonsensical instructions in both \vln pretraining and fine-tuning data to check how the performance of each model is affected.
We highlight our findings below.

\begin{table}[t]
\setlength{\aboverulesep}{0pt}
\setlength{\belowrulesep}{0pt}
\centering
\tabcolsep 7pt 
\begin{tabular}{>{\columncolor[gray]{0.8}}lrrrr}
Pretraining & \mlm & \mrm & \itm & \sap  \\
\toprule
{\small\orig} & $73.7$ & $31.9$ & $96.4$ & $73.1$ \\
{\small\sfword} & $49.7$ & $30.8$ & $94.9$ & $73.8$ \\
{\small\sfsent} & $74.9$ & $33.8$ & $95.7$ & $74.9$ \\
\midrule
{\small\permblock} & $73.0$ & $30.8$ & $37.3$ & $62.9$ \\
{\small\permall} & $72.9$ & $32.1$ & $36.2$ & $63.0$ \\
\midrule
{\small0\% language} & $4.0$ & $31.3$ & $36.5$ & $63.4$ \\
\bottomrule
\end{tabular}
\caption{
The accuracy of \hamt pretraining tasks on R2R validation unseen set. 
The \mlm task is most affected by word order noising, though that drop is not reflected in downstream success rate (Table~\ref{tab:shuffling}).
By contrast, all pretraining tasks except \mrm are affected by mismatch and language deletion.
}
\label{tab:proxy_acc}
\end{table}

\paragraph{Shuffling in pretraining does not affect SR.}
\sfword and \sfsent have almost has no negative impact on the SR (Table~\ref{tab:shuffling} column \orig) for either \hamt and \vlnb.
The standard deviation for \hamt is less than 1\% for all the experiments (Appendix~\ref{secapp:exps}).
\sfword reduces success rate less than \sfsent.
From that finding, we hypothesize that \vln agents' focus on object words~\cite{zhu-etal-2022-diagnosing} preserves performance since salient objects are kept in order across sentence boundaries in \sfword.
For noisier shuffles, \sfwordsent and \sfallword also show less than 3\% SR drop. 
We compare the pretraining proxy task performance in Table~\ref{tab:proxy_acc}.
Except for \mlm, all other three grounding tasks are not affected by the word shuffling, which indicates a weak language module might be sufficient for \vln fine-tuning.

\paragraph{Shuffling during fine-tuning slightly affects SR.}
Using un-noised pretraining, all shuffling methods on \hamt and \vlnb exhibit a 1.7-6.8\% drop in SR when applied at the fine-tuning stage (Table~\ref{tab:shuffling}).
However, even if both pretraining and fine-tuning data are shuffled, SR usually does not fall more than 10\%, indicating models can recover some kind of grounding for downstream, unperturbed test instructions from nonsensical training.
This counter-intuitive finding highlights the lack of word-order sensitivity in these SotA \vln agents.

\paragraph{High agreement with the original model.}
We observe high agreement of the above shuffling trained models \wrt the \orig. 
To get the variance of training, we test the agreement in the \orig setup across 3 different random seeds.
The pairwise \agreement are 0.88, 0.85, 0.86 for all 3 pairs. 
For models with shuffling in pretraining, \agreement with the \orig is 0.78, 0.79, 0.75 and 0.78 for \sfword, \sfsent, \sfwordsent and \sfallword, respectively.
The agreement is high, but it has a 6-13\% gap with the agreement across \orig seeds.
For models with shuffling only during fine-tuning, \agreement with the \orig is 0.84, 0.84, 0.78, and 0.78 for \sfword, \sfsent, \sfwordsent and \sfallword, respectively, nearly matching the un-noised \orig.

\begin{table}[t]
\setlength{\aboverulesep}{0pt}
\setlength{\belowrulesep}{0pt}
\centering
\begin{tabular}{>{\columncolor[gray]{0.8}}lrr}
Mismatch & \textbf{\hamt} & \textbf{\vlnb} \\
\toprule
No mismatch & $57.8$ & $54.7$ \\
\permblock & $49.0$  & $50.9$ \\
\permall & $51.8$ & $54.2$ \\ 
No pretraining & $45.8$ & $44.3$ \\
\bottomrule
\end{tabular}
\caption{The \sr on R2R validation unseen set when pretraining with irrelevant instructions. 
Irrelevant instructions are better than no pretraining at all.}
\label{tab:mismatch}
\end{table}
\begin{table}[t]
\setlength{\aboverulesep}{0pt}
\setlength{\belowrulesep}{0pt}
\centering
\begin{tabular}{>{\columncolor[gray]{0.8}}rrr}
\multicolumn{1}{>{\columncolor[gray]{0.8}}c}{\% Traj w\ Lang} & \textbf{\hamt} & \textbf{\vlnb} \\
\toprule
$100$ & $57.8$ & $54.7$ \\
$50$ & $56.0$ & $54.5$ \\
$1$ & $25.1$ & - \\
$0$ & $27.2$ & $28.7$ \\
\midrule
 No pretraining & $45.8$ & $44.3$ \\
\bottomrule
\end{tabular}
\caption{The \sr on R2R validation unseen set when pretraining with some portion of no-language trajectories. 
Zero or only a small portion of annotated trajectories can be harmful in pretraining, while using only half annotated with language is enough to recover the performance matching full annotations.}
\label{tab:nolang}
\end{table}

\begin{figure*}[t]
    \centering
    \includegraphics[width=0.95\textwidth]{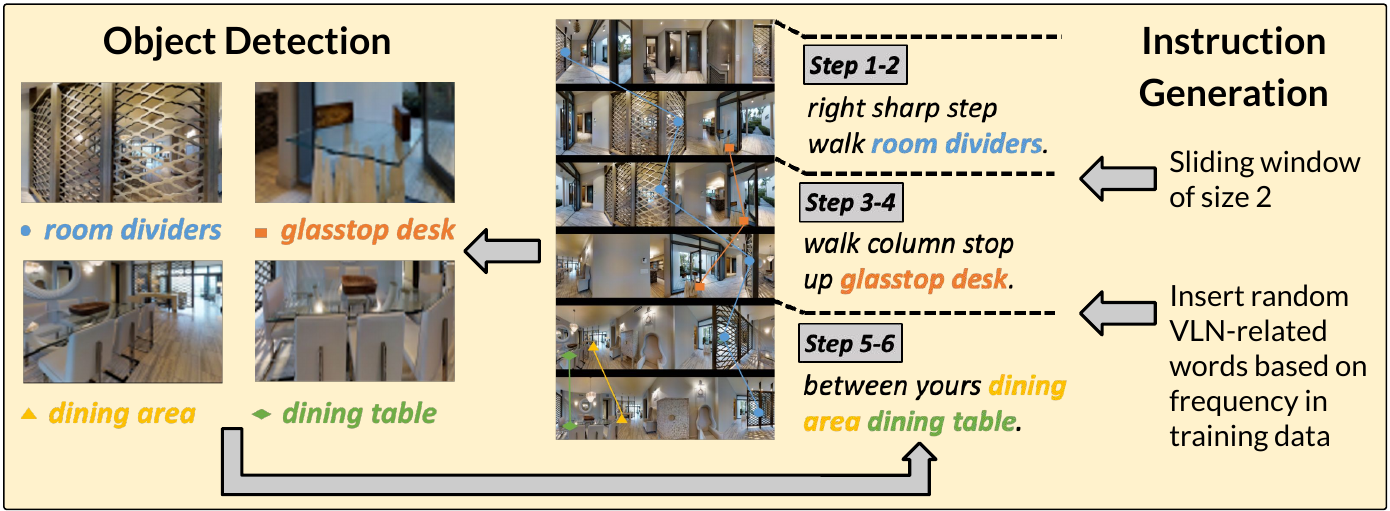}
    \caption{An illustration of object-based unigram sampling. We use an object detector along the trajectory and then generate instructions by inserting random words between the ordered object words.}
    \label{fig:unigram}
\end{figure*}

\subsection{\permutation}
Shuffling words within an instruction preserves word-image co-occurrences. 
Thus, one explanation for the results in Section~\ref{ssec:shuffling} is that \vln models only focus on the objects that appear in the trajectories and are mentioned in the instructions.

To break the word-image co-occurrence structure, we introduce mismatches between instruction annotations and trajectories.
Each trajectory has multiple (3-4) instruction annotations.
We explore \permblocklong (\permblock) and \permalllong (\permall).
The former assigns instructions to trajectories as blocks, such that after the assignment, all the instructions for one trajectory are describing the same (wrong) trajectory.
The latter assigns instructions randomly to trajectories, only keeping the number of instructions describing each trajectory the same between the assignment.
In both assignments, each instruction has meaningful language but mismatched grounding signals; in the former, there is consistency in the language descriptions of a trajectory.
We evaluate whether preserving the language-oriented proxy task signals and breaking the grounding-oriented proxy task signals will affect the VLN pretraining.

\paragraph{\permutation lowers SR, but still helps over no pretraining.}
Table~\ref{tab:proxy_acc} and~\ref{tab:mismatch} show mismatching the \vln trajectories hurts the performance of the \itm task, and affects the SR on the unseen room evaluation. However, the SR with mismatched instructions pretraining is still better than no pretraining.
In the \textsc{no pretraining} setting, a randomly initialized model is fine-tuned for the \vln task directly.
Higher randomness while mismatching helps to increase SR in both models (\permall > \permblock). The agreement with the \orig model is 0.72, 0.70 and 0.68 for \permall, \permblock and no pretraining, which drops in the same trend as the \sr.

\subsection{Empty Language Pretraining}
To test the importance of language during pretraining, we pretrain the proxy tasks with emptry strings for instructions for a portion of the R2R and augmented trajectories (\hamt) or just R2R trajectories (\vlnb).
Table~\ref{tab:nolang} summarizes these experiments.
Note that at 1\% of the data being paired with language annotations we run pretraining only for \hamt, since \vlnb pretraining uses only the human-annotated R2R data and 1\% is only a handful of trajectories (less than 50).

\paragraph{Some language is required in \vln pretraining}
Tables~\ref{tab:proxy_acc} and~\ref{tab:nolang} show that empty language pretraining can damage the language encoder beyond finetuning recovery, and reduce performance on the \itm and \sap tasks.
Pretraining with totally empty language data is worse than not pretraining at all for downstream task SR (Table \ref{tab:nolang}).
However, with only 50\% of trajectories annotated with language, both models have close performance to the \orig pretraining (Table~\ref{tab:nolang}).
For \hamt, the agreement with the standard model is 0.77, 0.57, 0.58 for 50\%, 1\%, and 0\% language pretraining, respectively, showing the \vln models with little language in pretraining have poor correlation with the full models.

\subsection{\ugtitle}
To test whether nonsensical data provides useful augmentation, we propose an efficient method to generate nonsensical instructions without a back-translated model, which improves the performance of downstream \vln over pretraining on training data alone ($53.7\rightarrow54.4$ SR).
Inspired by \sfsent in~\S\ref{ssec:shuffling}, we hypothesized that object-related words being in roughly the right order is important in grounding between instructions and trajectories.
To generate an instruction for a given unlabeled trajectory, we apply an object detector to each frame and concatenate some object words corresponding to detections together with randomly sampled filler words from a unigram model trained on the human-composed R2R instructions, which only requires CPU in the instruction generation time. 
We denote the method as \uglong (\ug; see Appendix~\ref{secapp:impl} for details).

\begin{table}[t]
\setlength{\aboverulesep}{0pt}
\setlength{\belowrulesep}{0pt}
\centering
\tabcolsep 3pt 
\begin{tabular}{>{\columncolor[gray]{0.8}}lrrr}
Pretraining data & \multicolumn{2}{c}{\textbf{\hamt}}  & \textbf{VLNBERT} \\
& IL & +RL & Scoring \\
\toprule
No pretraining & $45.8$ & $49.0$ & $44.3$ \\
\midrule
R2R trajs \\
{\phantom{-}w/ \orig} & $53.7$ & $58.5$ & $54.7$ \\
{\phantom{-}w/ \ug} & $43.7$ & $49.9$ & $54.5$ \\
\midrule
R2R \& Aug trajs \\
{\phantom{-}w/ \orig} & $57.8$ & $63.8$ & - \\ 
{\phantom{-}w/ \ug} & $51.0$ & $56.5$ & - \\
{\phantom{-}w/ \ug,\permall} & $54.4$ & $59.7$ & - \\
\bottomrule
\end{tabular}
\caption{The \sr on R2R validation unseen set when pretraining with annotations drawn from \uglong (\ug).
Only using noisy instructions from \ug works as well as human annotation (\orig) for \vlnb.
\hamt requires semantically-correct random instructions combined for \ug to be an effective augmentation.}
\label{tab:unigram}
\end{table}

\paragraph{\uglong is effective with pretrained language encoder.}
In Table~\ref{tab:unigram}, we replace R2R trajectory annotations with \ug generated ones, and pretraining with original R2R training data plus augmentation data from either a Speaker model (\orig) or \ug.
We observe different behaviors for \hamt and \vlnb with \ug. 
For \hamt, pretraining with \ug seems to be harmful, as using \trainonly with \ug is 2\% worse than no pretraining. On the other hand, \vlnb shows a large improvement with pretraining with \ug. 
As \vlnb is initialized with \vilb and \ug involves no meaningful language, we suspect \ug is useful when the language encoder is already pretrained.
Generating \ug data on augmented trajectories helps to eliminate the negative impact.
It shows 5.2\% better on \sr (Table~\ref{tab:unigram}) than the no pretraining baseline, which also verifies the quantity is more important.

\paragraph{Meaningful language is needed for randomly-initialized language encoders.}
To help the non-pretrained language encoder in \hamt, we add meaningful but arbitrary language instructions from \permall.
Specifically, for each trajectory, we assign three annotations from \ug and three annotations from \permall.
While only using \ug or \permall is inferior to \trainonly, mixing nonsensical and irrelevant instructions together outperforms the \trainonly baseline.
The \agreement with the original training also improves from 0.72 (\ug) to 0.74 (\ug,\permall) by mixing the two strategies.

\begin{table*}[t]
\setlength{\aboverulesep}{0pt}
\setlength{\belowrulesep}{0pt}
\centering
\begin{tabular}{l>{\columncolor[gray]{0.8}}r>{\columncolor[gray]{0.8}}r>{\columncolor[gray]{0.8}}r>{\columncolor[gray]{0.8}}r>{\columncolor[gray]{0.8}}rrrrrr}
Pretrain Shuffle$\rightarrow$ & \multicolumn{5}{>{\columncolor[gray]{0.8}}c}{\orig} & \multicolumn{5}{c}{\sfword} \\
Pretrain Data~$\downarrow$ & \sr & \mlm & \mrm & \itm & \sap & SR & \mlm & \mrm & \itm & \sap \\
\toprule
\trainandaug & $\pmb{57.8}$ & $75.4$ & $31.4$ & $\pmb{95.9}$ & $74.5$ & $\pmb{58.1}$ & $\pmb{49.6}$ & $30.8$ & $94.9$ & $73.8$ \\
\trainonly & $53.7$ &  $75.3$ & $27.8$ & $94.8$ &  $\pmb{74.6}$ & $51.7$ & $44.9$ & $27.2$
& $92.8$ & $72.1$ \\
\augonly & $57.7$ & $\pmb{75.7}$ & $\pmb{31.9}$ & $95.7$ & $74.6$ & $57.2$ & $49.2$ & $\pmb{31.7}$ & $\pmb{95.4}$ & $\pmb{74.0}$ \\
\augonlytrainsize & $53.5$ & $69.9$ & $27.8$ & $92.2$ & $67.3$ & $50.9$ & $44.5$ & $27.5$ & $93.1$ & $71.4$ \\
\bottomrule
\end{tabular}
\caption{The \sr on R2R validation unseen set and accuracy of proxy tasks in \hamt pretraining with original and word shuffled language.
The results suggest (1) data quantity matters more than quality, (2) augmentation primarily increases the accuracy of \mrm, and (3) augmentation improves the robustness of the models.}
\label{tab:qual_quan}
\end{table*}

\section{Quality \textit{vs.} Quantity}
\label{sec:qvq}

The previous section showed that the \emph{quality} of pretraining data matters surprisingly little.
Now we consider a complementary question: how much does \emph{quantity} of pretraining data matter, compared with quality?
Recall that \hamt was pretrained on the union of two datasets: the small, high-quality human-annotated R2R dataset, and a dataset of one million model-generated augmentation instructions that are known to be of much lower quality \cite{wang2021more}.\footnote{In Appendix~\ref{secapp:quality}, we apply dataset cartography~\cite{swayamdipta2020dataset} to these two sets of instructions to visualize this difference in quality.} 
To investigate the role of quantity in \vln pretraining, 
we pretrain \hamt with either only the R2R training data (\trainonly) or only the noisy augmentation data (\augonly), then fine-tune on R2R training data.

\paragraph{Quantity is more important than Quality.}
\augonly has roughly the same SR as pretraining on both \trainandaug data, and is 4\% better than the \trainonly (Table~\ref{tab:qual_quan}).
The \augonly model has 0.80 \agreement with the \trainandaug model, while the \trainonly model has 0.73 \agreement.
Combined training is more similar to \augonly training.

\paragraph{Quantity especially helps vision pretraining.} We also analyze the effect of changing the pretraining data on the pretraining proxy task objectives measured on held-out data.
The noisy augmentation data primarily improves the model's \mrm (+3\%) accuracy (an image-only proxy task), likely because it includes $38\times$ more visual trajectories than training data.
Meanwhile, even though the correspondence between instructions and trajectories is noisy, \mlm accuracy is not affected, as the language itself is still semantically meaningful. 
By increasing the quantity of low-quality data, \itm accuracy is even improved by 1\%.
We conclude quantity of \emph{visual trajectories} matters more than quality of instructions in \vln pretraining.
Thus, increasing the number of visual trajectories and pairing them with noisy language annotations is more helpful than cleaning the existing \vln data for \vln pretraining.
However, recall that having some language instructions is still crucial, since augmenting the visual trajectories with empty instructions is harmful to \vln pretraining (Table~\ref{tab:nolang}).

\paragraph{Augmentation improves robustness to shuffling.}
Pretraining with augmentation data also improves robustness to word shuffling. 
\sr on \trainonly has a 2\% SR drop whereas \augonly has only a 0.4\% point drop with \sfword in pretraining. 
For \sfword pretraining, using the augmentation data outperforms the \trainonly on all proxy tasks.

\begin{table}[t]
\setlength{\aboverulesep}{0pt}
\setlength{\belowrulesep}{0pt}
\centering
\tabcolsep 15pt 
\begin{tabular}{>{\columncolor[gray]{0.8}}lr}
Pretraining & \textbf{\hamt} \\
\toprule
{\orig} & $51.4$  \\  
{\sfword} & $46.7$\\  
{\sfsent} & $50.4$\\  
{\sfwordsent} &  $46.0$\\  
{\sfallword} & $41.3$\\ 
\midrule 
{\permall} & $45.4$ \\ 
\midrule
{\orig+\rxraug} & $52.9$ \\
\midrule
No pretraining & $34.7$ \\

\bottomrule
\end{tabular}
\caption{
The \sr on RxR validation unseen set when shuffling within instruction. The results show shuffling word order affects SR performance more on RxR, while irrelevant instructions work as well as in R2R.
}
\label{tab:noising_rxr}
\end{table}
\begin{table}[t]
\setlength{\aboverulesep}{0pt}
\setlength{\belowrulesep}{0pt}
\centering
\tabcolsep 4pt 
\begin{tabular}{>{\columncolor[gray]{0.8}}lrrrr}
Pretraining & te & hi & en-in & en-us \\
\toprule
{\orig} & $51.0$ & $53.0$ & $50.7$ & $49.0$ \\  
{\sfword} & $47.0$	& $48.0$ & $45.7$ & $43.5$\\  
{\sfsent} & $49.3$	& $52.1$ & $50.5$ & $47.9$\\  
{\sfwordsent} &  $45.4$ & $47.6$ & $45.6$ & $44.2$\\  
{\sfallword} & $40.9$ & $43.1$ & $40.8$ &$38.6$\\ 
\bottomrule
\end{tabular}
\caption{
The \sr breakdown RxR validation unseen set of different languages when shuffling within instruction.
}
\label{tab:language_breakdown}
\end{table}

\begin{figure}[t]
    \centering
    \includegraphics[width=\linewidth]{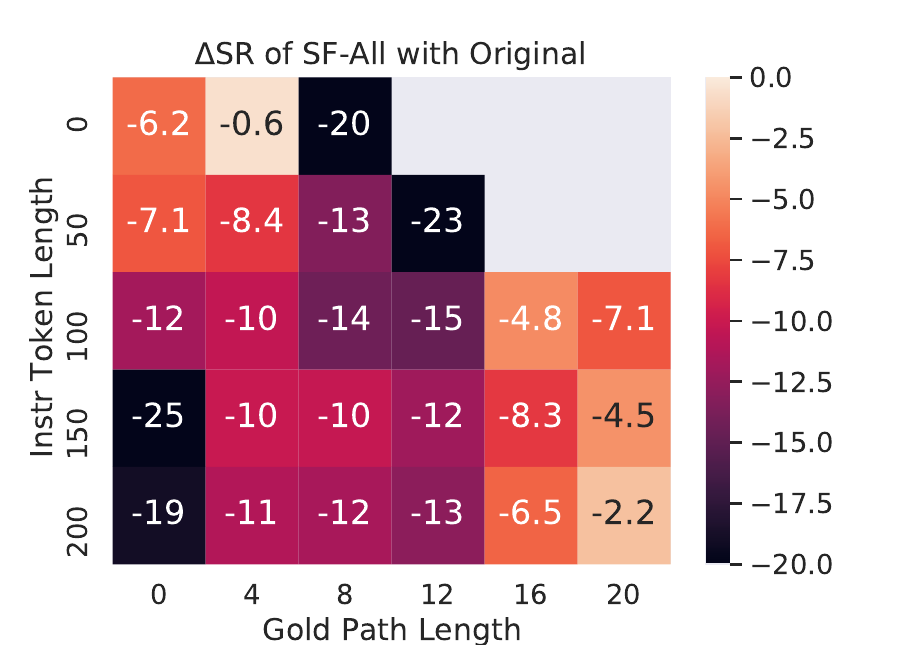}
    \caption{$\Delta$SR of \sfallword and \orig models over different length of instruction tokens (y-axis) and gold trajectories (x-axis). SR drops more on long instruction-short trajectory pairs, and short instruction-long trajectory pairs. }
    \label{fig:delta_sfall_rxr}
\end{figure}

\section{Analysis on RxR}
\label{sec:strategy_rxr}

We apply the same set of noising strategies to create nonsensical or irrelevant multilingual instructions on \rxr (RxR; \citet{rxr}), which has 11k trajectories and 79k multilingual human instructions as training data, and 1.5k trajectories and 14k multilingual human instructions as the validation data in unseen rooms. 
The instructions are in English, Hindi, or Telugu.
In the same simulated environment as R2R, RxR has a broader range of trajectory lengths.
We additionally consider the generated multilingual augmentation data from Marky-mT5~\cite{wang2021more}, which covers 368k trajectories and around 1 million multilingual instruction-trajectory pairs.
We use ``|'' and ``.'' as the separator for sentences in Hindi and Telugu, respectively.
We observe that nonsensical instructions affect SR more on RxR, while irrelevant instructions have a similar pattern compared to R2R.

\paragraph{Shuffling word order affects SR performance more on RxR.}
We pretrain with shuffled RxR data and finetune on the \orig RxR data. 
In Table~\ref{tab:noising_rxr}, shuffled data affects SR performance (up to $-10\%$) more than that on R2R (up to $-2\%$).

SR drops equally across difference languages. As shown in Table~\ref{tab:language_breakdown}, we break down SR on instructions in Telugu, Hindi, English annotated by Indian speaker, and English annotated by US speaker. 
Each language has similar SR drop on \sfword($\approx{-5}\%$), \sfsent($\approx{-1}\%$), \sfwordsent($\approx{-5}\%$) and \sfallword($\approx{-10}\%$).

From Fig.~\ref{fig:delta_sfall_rxr}, we compare the SR of \sfallword and \orig over different length of instruction tokens and trajectories.
We observe SR drops more on long instruction-short trajectory pairs, and short instruction-long trajectory pairs. 
When both instructions and trajectories are short, shuffling affects SR less.
When both instructions and trajectories are long, the absolute SR of both models are bad ($<10\%$).
We hypothesize that effects of noisy pretraining data between R2R and RxR are mainly attributable to the longer instructions and trajectories in RxR, where language guidance begins to play a larger role in model success.

\paragraph{Irrelevant instructions work similarly to R2R}
The \permall row in Table~\ref{tab:noising_rxr} shows that irrelevant instructions negatively affect SR, but not as much as not pretraining at all.


\section{Conclusion and Discussion}
\label{sec:conc}

In this paper, we analyze the importance of data quality and quantity for \vln pretraining.
We find that decreasing the quality of instructions by shuffling words or pairing them with unrelated trajectories still preserves many of the benefits of pretraining in the Room-2-Room benchmark. 
Increasing the quantity of visual trajectories in pretraining, with some corresponding noisy language annotation, is more helpful than increasing the quality of the existing VLN data. 
Inspired by these findings, we propose a simple, low-cost augmentation method that improves \vln pretraining performance, and could be used to annotate a large amount of new visual trajectories.

Our findings are related to the success of~\airb\cite{guhur2021airbert}, which uses an external source of visual trajectories with noisy annotations as augmentation. 
Based on our findings, scaling up the number of new trajectories and annotating with noisy augmentation is likely an effective method for improving transformer-based \vln pretraining.

Our work shows the current transformer-based \vln agents do not focus much on word ordering information. 
This counterintuitive behavior is suboptimal, and suggests the R2R benchmark may not require the detailed language grounding we hope, as a community, that \vln as a task demands.
Our results on the Room-across-Room benchmark, by contrast, suggest that longer trajectories that require more verbose language instructions may guard against models that can get by without substantial language reasoning.


\section*{Limitations}
We use R2R dataset for studying state-of-the-art VLN agents' behavior. 
The dataset consists of only English instructions and urban luxurious indoor scenes. 
We study the effect of word ordering within the instruction, and such order information maybe more or less sensitive in English language as compared to other languages. 
Our study is also limited to one VLN environment, i.e., Matterport3D, and the graphical VLN environment.
Therefore, this analysis may not be applicable to VLN datasets on other environments, such as continuous VLN environment~\cite{krantz_vlnce_2020} (which may allow more visual path augmentation), and outdoor VLN environments, such as Touchdown~\cite{chen2019touchdown}.

\bibliography{anthology, custom}
\bibliographystyle{acl_natbib}

\appendix

\clearpage
\begin{table*}[t]
\setlength{\aboverulesep}{0pt}
\setlength{\belowrulesep}{0pt}
\centering
\small
\tabcolsep 1.5pt 
\begin{tabular}{l>{\columncolor[gray]{0.8}}lrrrrrrrrrrrrrrr}
 \vln Model &  Pretraining Shuffle & \multicolumn{15}{c}{\cellcolor{blue!20} Finetuning Shuffle} \\
& & \multicolumn{3}{c}{\cellcolor{blue!20}{\orig}} & \multicolumn{3}{c}{\cellcolor{blue!20}{\sfword}} & \multicolumn{3}{c}{\cellcolor{blue!20}{\sfsent}} & \multicolumn{3}{c}{\cellcolor{blue!20}{\sfwordsent}} & \multicolumn{3}{c}{\cellcolor{blue!20}{\sfallword}} \\
& & \scriptsize\sr & \scriptsize\spl & \scriptsize\ndtw & \scriptsize\sr & \scriptsize\spl & \scriptsize\ndtw & \scriptsize\sr & \scriptsize\spl & \scriptsize\ndtw & \scriptsize\sr & \scriptsize\spl & \scriptsize\ndtw & \scriptsize\sr & \scriptsize\spl & \scriptsize\ndtw \\
\toprule
\multirow{5}{*}{\textbf{\hamt}} 
& {\orig} & $ 57.8$ & $ 55.1$ & $ \bf 67.2$ & $ 56.0$ & $ 53.4$ & $ 65.7$ & $ 54.6$ & $ 52.0$ & $ 65.0$ & $ 52.3$ & $ 49.7$ & $ 63.3$ & $ 53.1$ & $ 50.4$ & $ 64.0$ \\  
& {\sfword} & $ \bf 58.1$ & $ \bf 55.4$ & $ 66.8$ & $ \bf 56.6$ & $ \bf 53.9$ & $ \bf 65.8$ & $ 54.2$ & $ 51.7$ & $ 64.2$ & $ \bf 53.8$ & $ \bf 51.0$ & $ \bf 63.7$ & $ \bf 55.4$ & $ \bf 52.7$ & $ \bf 64.8$ \\  
& {\sfsent} & $ 56.3$ & $ 53.7$ & $ 66.1$ & $ 54.2$ & $ 51.3$ & $ 64.3$ & $ \bf 55.7$ & $ \bf 53.0$ & $ \bf 65.2$ & $ 53.3$ & $ 50.7$ & $ 63.7$ & $ 52.1$ & $ 49.4$ & $ 62.9$ \\  
& {\sfwordsent} & $ 54.9$ & $ 51.9$ & $ 65.1$ & $ 51.2$ & $ 48.6$ & $ 63.2$ & $ 50.5$ & $ 48.0$ & $ 62.0$ & $ 46.8$ & $ 44.3$ & $ 60.1$ & $ 48.0$ & $ 45.5$ & $ 60.1$ \\  
& {\sfallword} & $ 55.9$ & $ 53.2$ & $ 65.8$ & $ 52.6$ & $ 50.2$ & $ 63.4$ & $ 51.0$ & $ 48.5$ & $ 62.2$ & $ 49.8$ & $ 47.1$ & $ 61.4$ & $ 50.0$ & $ 47.5$ & $ 62.0$ \\  
\midrule
\multirow{5}{*}{\textbf{\vlnb}} 
& {\orig} & $ 54.7$ & $ 51.0$ & $ \bf 63.1$ & $ 53.5$ & $ 49.3$ & $ 61.6$ & $ 52.1$ & $ 48.6$ & $ 62.0$ & $ 45.5$ & $ 42.3$ & $ 58.7$ & $ 47.6$ & $ 44.2$ & $ 58.5$ \\  
& {\sfword} & $ \bf 55.8$ & $ \bf 51.2$ & $ 61.7$ & $ \bf 55.9$ & $ \bf 51.6$ & $ \bf 62.1$ & $ 52.6$ & $ 48.7$ & $ 61.5$ & $ 47.0$ & $ 43.1$ & $ 58.4$ & $ 51.0$ & $ 46.8$ & $ 60.0$ \\  
& {\sfsent} & $ 54.7$ & $ 51.0$ & $ 62.3$ & $ 52.4$ & $ 48.0$ & $ 60.8$ & $ \bf 53.3$ & $ \bf 49.5$ & $ \bf 62.1$ & $ 49.5$ & $ 46.2$ & $ \bf 61.0$ & $ \bf 51.0$ & $ \bf 47.1$ & $ \bf 61.2$ \\  
& {\sfwordsent} & $ 53.0$ & $ 49.0$ & $ 61.1$ & $ 53.3$ & $ 48.9$ & $ 60.5$ & $ 52.8$ & $ 48.6$ & $ 61.3$ & $ \bf 50.6$ & $ \bf 46.9$ & $ 60.8$ & $ 28.3$ & $ 25.7$ & $ 44.8$ \\  
& {\sfallword} & $ 54.1$ & $ 50.2$ & $ 62.4$ & $ 54.8$ & $ 50.5$ & $ 61.7$ & $ 51.7$ & $ 47.2$ & $ 60.5$ & $ 48.5$ & $ 44.9$ & $ 59.0$ & $ 48.3$ & $ 44.4$ & $ 59.1$ \\
\bottomrule
\end{tabular}
\caption{
Under different pretraining and fintuning shuffling setup, the \sr, \spl, \ndtw results on R2R validation unseen set.
\hamt is finetuned with IL and averaged over 3 runs. \vlnb has one run.
}
\label{tab:shuffling_long}
\end{table*}
\begin{table*}[t]
\setlength{\aboverulesep}{0pt}
\setlength{\belowrulesep}{0pt}
\centering
\small
\tabcolsep 3pt 
\begin{tabular}{l>{\columncolor[gray]{0.8}}lrrrrrrrrrrrrrrr}
 \vln Model &  Pretraining Shuffle & \multicolumn{15}{c}{\cellcolor{blue!20} Finetuning Shuffle} \\
& & \multicolumn{3}{c}{\cellcolor{blue!20}{\orig}} & \multicolumn{3}{c}{\cellcolor{blue!20}{\sfword}} & \multicolumn{3}{c}{\cellcolor{blue!20}{\sfsent}} & \multicolumn{3}{c}{\cellcolor{blue!20}{\sfwordsent}} & \multicolumn{3}{c}{\cellcolor{blue!20}{\sfallword}} \\
& & \scriptsize\sr & \scriptsize\spl & \scriptsize\ndtw & \scriptsize\sr & \scriptsize\spl & \scriptsize\ndtw & \scriptsize\sr & \scriptsize\spl & \scriptsize\ndtw & \scriptsize\sr & \scriptsize\spl & \scriptsize\ndtw & \scriptsize\sr & \scriptsize\spl & \scriptsize\ndtw \\
\toprule
\multirow{5}{*}{\textbf{\hamt}} 
& {\small\orig} & $ 0.3$ & $ 0.3$ & $ 0.1$ & $ 0.1$ & $ 0.2$ & $ 0.0$ & $ 0.2$ & $ 0.2$ & $ 0.2$ & $ 0.3$ & $ 0.3$ & $ 0.2$ & $ 0.3$ & $ 0.4$ & $ 0.4$ \\  
& {\small\sfword} & $ 0.5$ & $ 0.6$ & $ 0.5$ & $ 0.1$ & $ 0.2$ & $ 0.4$ & $ 0.3$ & $ 0.2$ & $ 0.3$ & $ 0.4$ & $ \bf 0.4$ & $ 0.1$ & $ 0.5$ & $ 0.3$ & $ 0.3$ \\  
& {\small\sfsent} & $ \bf 0.7$ & $ \bf 0.6$ & $ 0.3$ & $ \bf 0.5$ & $ \bf 0.6$ & $ 0.2$ & $ \bf 0.8$ & $ \bf 0.8$ & $ \bf 0.6$ & $ 0.1$ & $ 0.2$ & $ 0.1$ & $ 0.4$ & $ 0.5$ & $ 0.3$ \\  
& {\small\sfwordsent} & $ 0.3$ & $ 0.4$ & $ \bf 0.7$ & $ 0.2$ & $ 0.2$ & $ 0.1$ & $ 0.6$ & $ 0.4$ & $ 0.3$ & $ 0.4$ & $ 0.4$ & $ \bf 0.4$ & $ \bf 0.7$ & $ \bf 0.7$ & $ \bf 0.5$ \\  
& {\small\sfallword} & $ 0.3$ & $ 0.1$ & $ 0.1$ & $ 0.4$ & $ 0.3$ & $ \bf 0.5$ & $ 0.6$ & $ 0.7$ & $ 0.4$ & $ \bf 0.5$ & $ 0.3$ & $ 0.1$ & $ 0.4$ & $ 0.4$ & $ 0.3$ \\  
\bottomrule
\end{tabular}
\caption{
Under different pretraining and fintuning shuffling setup, the standard deviation of \sr, \spl, \ndtw results on R2R validation unseen set over 3 runs.
}
\label{tab:shuffling_dev}
\end{table*}
\begin{table}[t]
\small
\setlength{\aboverulesep}{0pt}
\setlength{\belowrulesep}{0pt}
\centering
\tabcolsep 3pt 
\begin{tabular}{>{\columncolor[gray]{0.8}}lrrrrrr}
Mismatch & \multicolumn{3}{c}{\textbf{\hamt}} & \multicolumn{3}{c}{\textbf{\vlnb}} \\
& \small\sr & \small\spl & \small\ndtw & \small\sr & \small\spl & \small\ndtw \\
\toprule
{No mismatch} & $57.8$ & $55.1$ & $67.2$ & $57.8$ & $51.0$ & $63.1$\\
\permblock & $ 49.0$ & $ 46.6$ & $ 61.1$ &$54.2$ &$50.0$ & $61.2$\\
\permall & $ 51.8$ & $ 49.2$ & $ 63.3$ & $ 50.9$ & $ 47.5$ & $ 60.3$\\
No pretraining & $45.8$ & $43.5$ & $59.5$ & $44.3$ & $39.8$ & $54.3$ \\
\bottomrule
\end{tabular}
\caption{The \sr, \spl and \ndtw on R2R validation unseen set when pretraining with irrelevant instructions. Irrelevant instructions affect the performance, while still better than no pretraining.}
\label{tab:mismatch_long}
\end{table}
\begin{table}[t]
\small
\setlength{\aboverulesep}{0pt}
\setlength{\belowrulesep}{0pt}
\centering
\tabcolsep 3pt 
\begin{tabular}{>{\columncolor[gray]{0.8}}rrrrrrr}
\% Traj w\ Lang &\multicolumn{3}{c}{\textbf{\hamt}} & \multicolumn{3}{c}{\textbf{\vlnb}} \\
& \small\sr & \small\spl & \small\ndtw & \small\sr & \small\spl & \small\ndtw \\
\toprule
$100$ & $57.8$ & $55.1$ & $67.2$ & $57.8$ & $51.0$ & $63.1$\\
$50$ & $ 56.0$ & $ 53.5$ & $ 65.6$ & $54.5$ & $50.1$ & $61.3$\\  
$1$ & $ 25.1$ & $ 23.8$ & $ 44.0$ & - & - & - \\
$0$ & $ 27.2$ & $ 25.6$ & $ 44.8$ & $28.7$ & $26.3$ & $45.8$ \\  
\midrule
 No pretraining & $45.8$ & $43.5$ & $59.5$ & $44.3$ & $39.8$ & $54.3$ \\ 
\bottomrule
\end{tabular}
\caption{The \sr, \spl and \ndtw on R2R validation unseen set when pretraining with some portion of no-language trajectories. $^*$ refers to no augmented language. Little amount of trajectories with language can be harmful in pretraining, while half of them with language is enough to recover the performance.}
\label{tab:nolang_long}
\end{table}

\begin{table*}[t]
\setlength{\aboverulesep}{0pt}
\setlength{\belowrulesep}{0pt}
\small
\centering
\tabcolsep 6pt 
\begin{tabular}{>{\columncolor[gray]{0.8}}lrrrrrrrrr}
Pretraining data & \multicolumn{6}{c}{\textbf{\hamt}}  & \multicolumn{3}{c}{\textbf{\vlnb}} \\
& \multicolumn{3}{c}{IL} & \multicolumn{3}{c}{+RL} & \multicolumn{3}{c}{Scoring} \\
& \small\sr & \small\spl & \small\ndtw & \small\sr & \small\spl & \small\ndtw  & \small\sr & \small\spl & \small\ndtw  \\
\toprule
No pretraining & $45.8$ & $43.5$ & $59.5$ & $49.0$ & $46.9$ & $62.3$ & $44.3$ & $39.8$ & $54.3$  \\
R2R trajs  & $ $ & $ $ & $ $ $ $ & $ $ & $ $ $ $ & $ $ & $ $\\
{\small-- w/ \orig} & $ 53.7$ & $ 51.1$ & $ 63.1$ &$58.6$ & $55.0$ & $65.3$ & $57.8$ & $51.0$ & $63.1$\\ 
{\small-- w/ \ug} & $43.7$ & $41.4$ & $58.0$ &$49.9$ & $46.3$ & $60.4$ &$54.5$ & $50.3$ & $61.0$ \\
R2R \& Ang trajs \\
{\small-- w/ \orig} & $57.8$ & $55.1$ & $67.2$  &$63.8$ & $59.6$ & $68.7$ & - & - & -\\ 
{\small-- w/ \orig} & $57.8$ & $55.1$ & $67.2$  &$63.8$ & $59.6$ & $68.7$ & - & - & -\\ 
{\small-- w/ \ug} & $ 51.0$ & $ 48.4$ & $ 62.9$ &$56.5$ & $52.3$ & $63.8$ & - & - & - \\
{\small-- w/ \ug,\permall} & $ 54.4$ & $ 51.9$ & $ 65.1$& $59.7$ & $55.5$ & $66.0$ & - & - & -\\
\bottomrule
\end{tabular}
\caption{The \sr, \spl and \ndtw on R2R validation unseen set when pretraining with landmark-based unigram sampling. \ug: \uglong. Only using noisy instructions from \ug works as well as human annotation in \vlnb. \hamt requires semantically-correct random instructions combined for \ug to be an effective augmentation.}
\label{tab:unigram_long}
\end{table*}

\section{Complete Experiment Results}
\label{secapp:exps}

\subsection{Other Metrics and Standard Deviations}
\label{secapp:other_metrics}

For completeness, we replicate the tables in the main paper and add the \spl and \ndtw results here. We also report the agreement score and the loss of proxy tasks of \vlnb models.

\spl and \ndtw in Table~\ref{tab:shuffling_long} shows the similar trend as \sr, strengthening the findings that (1) increase the shuffling noise will decrease \vln performance, and (2) only severe noising by shuffling in pretraining starts having effects in final performance.
Through \sfword has higher average performance than the \orig training, the improvement is not significant, as the improvement across different measure is all within the standard deviation listed in Table~\ref{tab:shuffling_dev}.
Similarly, Table~\ref{tab:mismatch_long},~\ref{tab:nolang_long} and~\ref{tab:unigram_long} shows similar trend on \spl and \ndtw as on \sr.

We only show the agreement score of \hamt models in the main paper.
Additionally, we report the agreement score of \vlnb models here. 
Taking the original pretrained models as the model to compare to, \agreement on \sfword, \sfsent, \sfwordsent and \sfallword in pretraining is 0.71, 0.71, 0.71, 0.72, correspondingly. 
\agreement on \sfword, \sfsent, \sfwordsent and \sfallword in fine-tuning is 0.72, 0.69, 0.66, 0.66, correspondingly.
Different from \hamt, this could imply pretraining might not be the dominant stage with only a small number of pretraining proxy tasks.
\agreement on \permall, \permblock and no pretraining is 0.70, 0.68 and 0.65, which shows similar drops as \agreement in \hamt.
Likewise, \agreement on no language and half language training is 0.56, 0.69.
\agreement on \uglong is 0.68.

We report the loss of \mlm and \mrm tasks after \vlnb pretraining on its training set (\ie, the R2R training set), to show the difficulty of converging for each noising method.
Surprisingly, the mismatched trajectory training is almost as difficult as the original training, indicating \vlnb focus less on grounding between language and vision, and more on \mlm and \mrm as a single-modality task.

\begin{table}[t]
\setlength{\aboverulesep}{0pt}
\setlength{\belowrulesep}{0pt}
\centering
\tabcolsep 8pt 
\begin{tabular}{>{\columncolor[gray]{0.8}}lrr}
\ug Objects & \textbf{\hamt} & \textbf{\vlnb} \\
\toprule
In order & $\bf 51.0$ & $\bf 54.5$ \\
No order & $49.0$ & $52.2$ \\
No objects & $44.3$ & $50.3$ \\ 
\bottomrule
\end{tabular}
\caption{Ablations on the order and existence of landmark words in \ug. Out of order object words and no object words will be harmful.}
\label{tab:unigram_ablation}
\end{table}
\begin{table}[t]
\setlength{\aboverulesep}{0pt}
\setlength{\belowrulesep}{0pt}
\centering
\tabcolsep 5pt 
\begin{tabular}{>{\columncolor[gray]{0.8}}lrrrr}
\ug Objects & \mlm & \mrm & \itm & \sap  \\
\toprule
{\small\orig} & $73.7$ & $31.9$ & $96.4$ & $73.1$ \\
{\small\ug} \\
{\small\phantom{--}In order} & $61.9$ & $32.7$ & $81.8$ & $67.5$\\
{\small\phantom{--}No order} & $62.2$ & $33.3$ & $74.9$ & $68.1$ \\
{\small\phantom{--}No objects} & $47.9$ & $33.0$ & $35.6$ & $63.6$ \\
\bottomrule
\end{tabular}
\caption{The accuracy of \hamt pretraining tasks for different \ug variants on the R2R validation unseen set. \ug in order has no large gap with \orig on pretraining tasks. Removing the object detection module hurts \mlm and \itm.}
\label{tab:proxy_acc_unigram}
\end{table}

\subsection{Ablations on Objects in Sampling}
\label{secapp:uo_ablation}

To test the importance of ordered and relevant object words, we perform two ablations. 
First, we shuffle all the words in the generated instructions by \ug, such that the object words are out of the order as appeared in the trajectory. 
Second, instead of using an object detector, we include object words in the unigram model and randomly sample them.
Results in Table~\ref{tab:unigram_ablation} indicate that shuffling object words and removing detected object are both harmful to \vln performance.
Shuffling object order leads to similar \sr drops as shuffling all words in  Table~\ref{tab:shuffling}.
Removing the object detection module is even more harmful, showing that having some image-text correspondence does matter for \vln pretraining.
Removing detect objects reduces the performance in several pretraining tasks, such as \mlm and \itm, as in Table~\ref{tab:proxy_acc_unigram}.

\begin{table}[t]
\setlength{\aboverulesep}{0pt}
\setlength{\belowrulesep}{0pt}
\centering
\tabcolsep 10pt 
\begin{tabular}{>{\columncolor[gray]{0.8}}lrr}
Pretraining & \mlm & \mrm \\
\toprule
{\orig} & $1.35{e}^{-2}$ & $1.89{e}^{-2}$ \\
{\sfword} & $7.06{e}^{-2}$ & $2.04{e}^{-2}$  \\
{\sfsent} & $1.97{e}^{-2}$ & $1.94{e}^{-2}$ \\
\midrule
{\permall} & $1.45{e}^{-2}$ & $1.85{e}^{-2}$ \\
\midrule
{50\% language} & $7.13{e}^{-2}$ & $2.53{e}^{-2}$ \\
{0\% language} & $9.26{e}^{-6}$ & $1.75{e}^{-2}$ \\
\midrule
\ug \\
-- In order & $6.21{e}^{-2}$ & $2.25{e}^{-2}$ \\
-- No order & $4.87{e}^{-2}$ & $2.22{e}^{-2}$ \\
-- No objects & $4.09{e}^{-2}$ & $2.04{e}^{-2}$ \\
\bottomrule
\end{tabular}
\caption{
The loss of \vlnb pretraining tasks, \mlm and \mrm, on its training set under different noising setups.  
}
\label{tab:proxy_loss}
\end{table}

\section{Implementation Details}
\label{secapp:impl}

We report the implementation details in this section, including the details for \uglong sampling, the hyperparameters for \hamt and \vlnb pretraining and fine-tuning, and scientific artifacts we used.

\subsection{Details on \uglong (\ug)}

We apply a weakly supervised object detector from~\citet{zhu-etal-2020-babywalk} to generate object words for panoramas in the trajectory.
We train a unigram model by counting the frequency of word occurrence in the R2R training instructions.
To avoid redundancy, we remove object words (i.e., words that are labels in the object detector) from the unigram model. 
From the training data, we also model the distribution of the instruction length, \textit{i.e.}, the number of words in each instruction.  

When we sample a new instruction given a trajectory, we first sample the target instruction length from the distribution of instruction lengths.
As in Fig.~\ref{fig:unigram}, we generate top-$k$ objects from each panorama along the trajectory in order.
Then, starting from the first panorama, using a sliding window of size $b$, we sample $a$ objects among the detected $k\times b$ objects, and use these object words in our generated instructions.
In our experiments, we use $a=3, b=2, k=5$.
If the total object words have a word count less than the instruction length, we will insert words sampled from the unigram model between object words to create the whole instruction.

We sample and set the instruction length before generating it. 
If the total selected object words have a word count less than the instruction length, we will insert  words sampled from the unigram model to increase the instruction length to what we set.
We fill in each gap between object words with the same number of random words.
Suppose the trajectory length is $L$, we set the instruction length as $l$. 
As the sliding window size is $a$, we have $\lceil L/a \rceil$ sliding windows in total, each with $b$ objects.
We concatenate $b$ object in the sliding window and insert random words in front of each object words block.
Notice that one object could have multiple words, the total object word count could be more than $b\lceil L/a \rceil$.
Assuming we have $b^\star\lceil L/a \rceil$ object words in total, we insert $(l-b^\star\lceil L/a \rceil)/a$ words in front of each object word block, and put a period at the end to create a sentence to describe the sliding window.
We get the final instruction by concatenating all the sentences.

On the same machine with 1 Quadro RTX 6000 GPU, the training time for speaker mode and UO classifier is 32h and 2h, respectively. 
The \ug classifier requires an additional 0.5h to extract the object from all images. After that, the generation time per new data for speaker model and \ug is 0.02s and 0.01s, respectively. 

\subsection{\hamt setup}

\hamt use R2R train data \citet{anderson2018vision} and the synthesized augmented data. R2R includes 90 houses and 10,567 panoramas. The whole dataset is split into train, val seen, val unseen and test with 61, 59, 56, 11 scans respectively. The synthesized augmented data contains 178,270 trajectories and 1,069,620 instructions, and is only used in pretraining.

In pretraining, unlike \hamt, we do not perform the second-stage end-to-end training on ViT features for simplicity. We use AdamW optimizer with a weight decay of 0.01. The training and validation batch size are set as 64. We train for 200k steps with 10k warmup steps and a learning rate of $5{e}^{-5}$ on NVIDIA A100 GPU for 11 hours. The total pretraining cost 363 GPU hours on NVIDIA A100.

Following the {\hamt}~\cite{chen2021hamt} paper, we set $9$ layers for language transformer encoder, $9$ layers for panoramic transformer in history encoder, and $4$ layers or cross-modal transformer. During fine-tuning, we freeze the parameters of history and observation encoders. For IL, We use pretrained ViT end-to-end training feature and maximum of 10k iterations with a learning rate of $1{e}^{-5}$ on Quadro RTX 6000 for 1.5 hours for each run. For IL + RL, we use the same vision feature and train for a maximum of 25k iterations on Quadro RTX 6000 for 10 hours for each run. For both IL and IL + RL experiments, We evaluate on validation unseen set every 500 iterations. The batch size is 16 and we use an AdamW optimizer. The rest hyperparameters are the same as the \hamt model. The total fine-tuning cost 157.5 GPU hours on Quadro RTX 6000.

\subsection{\vlnb setup}
In pretraining of \vlnb, we train a total of 50 epochs for pretraining with a batch size of 64. We use the AdamW optimizer with a weight decay of 0.01. The initial learning rate is set as $4e^{-5}$, and a cooldown factor of 8 is used to decrease the learning rate through epochs.
One run of pretraining on 2 NVIDIA V100 GPUs takes 70 hours. The total pretraining cost 2100 GPU hours on NVIDIA V100.

In fine-tuning of \vlnb, we train a total of 20 epochs with a batch size of 64. We use a fixed learning rate of $4e^{-5}$ with the AdamW optimizer with a weight decay of 0.01. The rest hyperparameters are the same as the original \vlnb model. One run of fine-tuning on 2 NVIDIA V100 GPUs takes 24 hours.  The total fine-tuning cost 3600 GPU hours on NVIDIA V100.

\subsection{Scientific Artifacts and Licenses}
Our implementation is built on top of the publicly available codebase of \hamt~\cite{chen2021hamt} and \vilb~\cite{majumdar2020improving}. 
\hamt is under MIT license and \vlnb has no license in the codebase.
However, \vlnb claims non-exclusive license to distribute in the paper, and we only use their codebase for research purpose.

We use the dataset from~\citet{anderson2018vision}, which is under MIT license. The instructions in the dataset are in English. They are collected for navigation in a simulated environment. Based on our word tokenizer, the instructions contain no information that names or uniquely identifies individual people, or other offensive content.

\begin{figure*}
     \centering
     \begin{subfigure}[b]{0.45\textwidth}
         \centering
         \includegraphics[width=\textwidth]{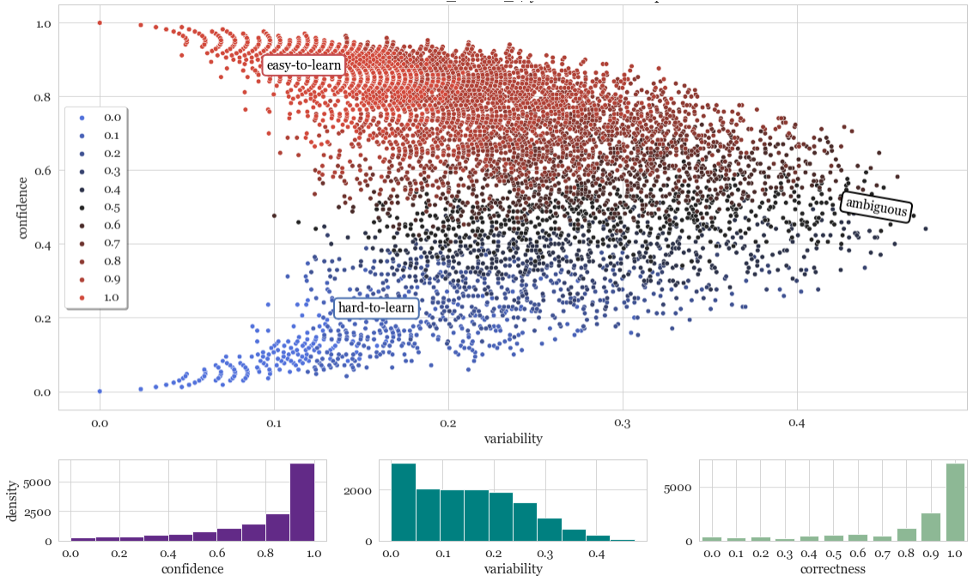}
         \caption{R2R data at the last epoch}
         \label{fig:dc_r2r}
     \end{subfigure}
     \hfill
     \begin{subfigure}[b]{0.49\textwidth}
         \centering
         \includegraphics[width=\textwidth]{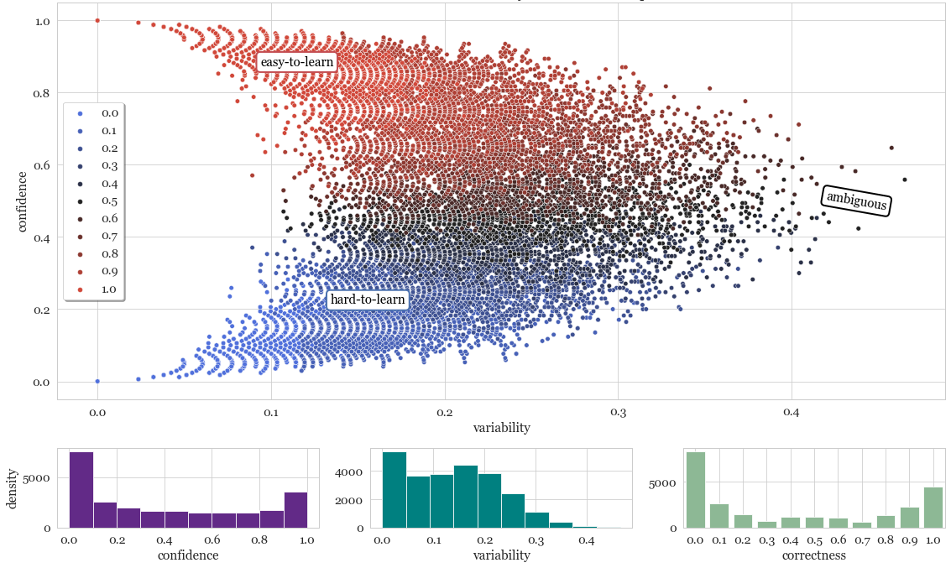}
         \caption{Synthesized data at the last epoch}
         \label{fig:dc_aug}
     \end{subfigure}
        \caption{VLN Data Cartography on (a) the R2R data and (b) the synthesized data.}   
\end{figure*}

\section{Data Quality Visualization through Cartography}
\label{secapp:quality}

\subsection{\dc}
\dc~\cite{swayamdipta2020dataset} is an automated visualization and analysis method for large datasets. It closely studies training dynamics, \ie, evolution of the model behavior over training epochs, using information such model confidence and variability. For a model trained on the dataset, $D = \{(\boldsymbol{x}, y^*)_i\}_{i=1}^N$, of size $N$, the confidence ($\hat{\mu}_i$) and variability ($\hat{\sigma}_i$) for $i$-th data point are computed as,

\begin{align}
\hat{\mu}_i&=\frac{1}{E} \sum_{e=1}^E p_{\boldsymbol{\theta}^{(e)}}\left(y_i^* \mid \boldsymbol{x}_i\right) \label{eq:conf} \\
\hat{\sigma}_i&=\sqrt{\frac{\sum_{e=1}^E\left(p_{\boldsymbol{\theta}^{(e)}}\left(y_i^* \mid \boldsymbol{x}_i\right)-\hat{\mu}_i\right)^2}{E}} \label{eq:var}
\end{align}

where $p_{\boldsymbol{\theta}^{(e)}}$ is the model's probability to predict the ground truth label $y_i^*$ given observation $\boldsymbol{x}_i$, with parameters $\theta^{(e)}$ for the training epoch $e$. $E$ is the total number of training epochs. 

\dc has been applied to study datasets in NLP. The data maps generated using this tool consists of three data regions: 1) \textit{ambiguous} data points characterized by high model variance, 2) \textit{easy-to-learn} points with low variance and high confidence, and 3) \textit{hard-to-learn} points with low variance and low confidence. 
The data regions and the plotting of confidence and variability, could be used to characterize the difficulty and noisiness in the dataset.
Inspired by this, we apply \dc to \vln to visualize and compare the quality of human-annotated and synthesized data.

\subsection{\dc Approach for \vln}
The \vln task requires taking several decisions over the time steps to complete the given instruction. Not as straight forward as classification, we have to define the model's probability $p_{\boldsymbol{\theta}^{(e)}}$ to predict the ground truth trajectory. 
To keep the computation efficient, we compute the probability of the ground truth trajectory by multiplying the probability of correct action at each step while following the instruction. 
Specifically, for a given instruction $X_i$ and a ground truth trajectory $Y_i=(y_i^1, ..., y_i^m)$ of length $m$, we have
\begin{align}
    p_{\boldsymbol{\theta}^{(e)}}(Y_i|X_i)=\prod_{j=1}^{m-1}p_{\boldsymbol{\theta}^{(e)}}(y_i^j|X_i, y_i^{1..j-1}) \label{eq:vlnprob}
\end{align}
This definition may change the scale of the probabilities, i.e., tending to lower values. 
In this scenario, it is important that the path-instruction alignment is correct, otherwise even if the agent reaches the correct goal but makes one bad mistake along the path, the probability will be very low. 
Mathematically, however, the scale will be preserved with multiplied path probabilities.
Thus, in theory, it may still be a valid method to compute data maps. Even though the data maps will be more concentrated towards the low probability regions, but hopefully the maps still represent the inherent difficulty of the task in \vln, where the agent indeed needs to make multiple correct decisions to complete the task as opposed to a single prediction.
We train the model using teacher forcing, so that the agent in training is forced to follow the correct path. This enables the computation of the model's probability for the correct direction/action prediction at any time step. Using the overall trajectory probability, we compute model confidence and variability from Eqns.~(\ref{eq:conf},~\ref{eq:var},~\ref{eq:vlnprob}). 
We use these values to identify the data regions, namely ambiguous, easy-to-learn, and hard-to-learn. 

For the R2R training data, we train \hamt on R2R for 20 epochs and  plot the cartography on the data. For the speaker-augmented data, likewise, we save the models trained on R2R for each epoch, and then use them to evaluate the augmentation data. From Fig.~\ref{fig:dc_aug}, we observe that the R2R examples are mostly in the easy-to-learn region, whereas the augmentation examples are mostly in the ambiguous and hard-to-learn regions.
We could see it more clearly on the small plot in purple under the cartography plot, in which the distribution of confidence on R2R is lean towards 1.0, and the distribution of confidence on the augmentation data is lean towards 0.0. 
The cartography plots verify the fact that the augmentation data is quantitatively more noisy than the R2R training data.


\section{Selected Data Training through Cartography}
\label{secapp:remove}
Given our findings on quantity matters more than quality on \vln pretraining, in the way that reducing the augmentation data quality even more does not hurt the performance. A natural question to ask is, on the other hand, how will the \vln performance change if we increase the data quality?

We focus fine-tuning, and we experiment on two setups, (1) fine-tuning \hamt after pretraining. (2) fine-tuning \hamt from scratch without any pretraining.
Apart from R2R, we also experiment with the RxR dataset.
Similar to \citet{swayamdipta2020dataset}, we perform data editing by using or removing different regions of the data for fine-tuning the \hamt model, trying to explore if the following findings from~\citet{swayamdipta2020dataset} still hold under in the \vln setup: ambiguous data helps in generalization, easy-to-learn data aids convergence, and hard-to-learn data are mostly annotation errors. 
Besides, we apply the selected data training to understand if improving data quality by identifying useful data regions will improve \vln agents.

\begin{figure*}
     \centering
     \begin{subfigure}[b]{0.45\textwidth}
         \centering
         \includegraphics[width=\textwidth]{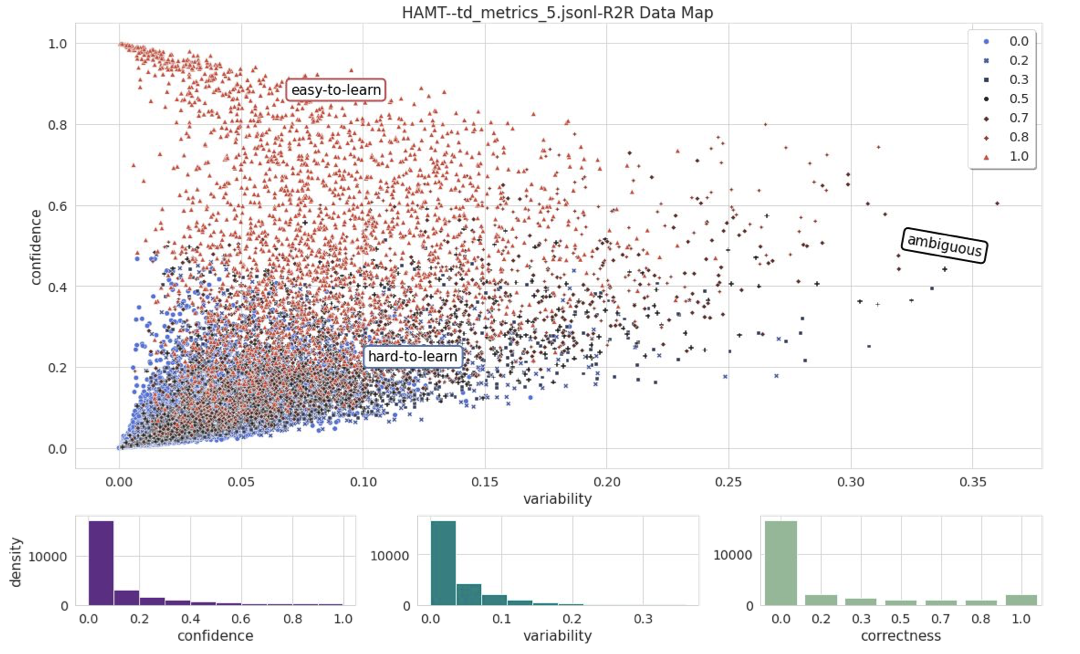}
         \caption{At epoch 5 (best validation performance)}
         \label{fig:dc_rxr}
     \end{subfigure}
     \hfill
     \begin{subfigure}[b]{0.49\textwidth}
         \centering
         \includegraphics[width=\textwidth]{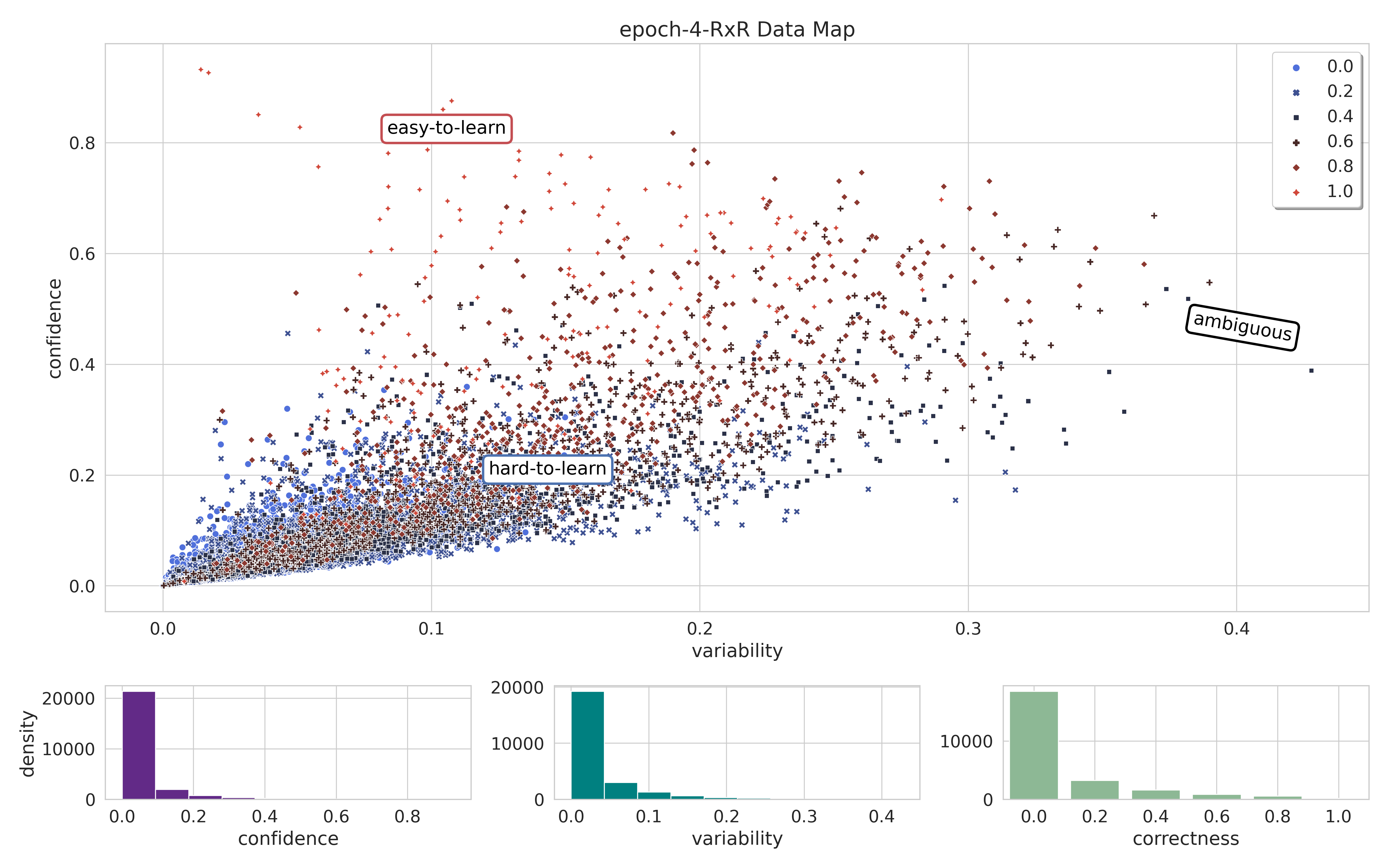}
         \caption{At epoch 4 (best validation performance)}
         \label{fig:dc_rxr_wo_pt}
     \end{subfigure}
        \caption{VLN Data Cartography on RxR data: (a) with pretraining, (b) without pretraining}
\end{figure*}

\begin{table*}[ht]
\centering
\caption{Results for RxR data editing for training HAMT model from scratch and initializing from pretrained HAMT. We use overall path probability to compute confidence and variability, based on the best epoch of the model trained on full data without pretraining. ``random 0.x'' refers to training on 0.x fraction of data. ``cut\_amb 0.x'' refers to removing 0.x fraction of data with the highest variability, "top\_amb 0.x" refers to training on 0.x fraction of data with the highest variability. Similarly, ``top\_conf 0.x'' refers to training on 0.x fraction of data with the highest confidence. \textbf{Bold} represents all results, better than full data training. `-' are runs yet to complete.}
\label{tab:rxr_remove}
\tabcolsep 12pt
\begin{tabular}{lrrrrrr}
\textbf{Training Data Fraction}  & \multicolumn{3}{c}{\textbf{No pretraining}} & \multicolumn{3}{c}{\textbf{With pretraining}} \\
\cmidrule(lr){2-4}\cmidrule(lr){5-7}
 & \sr & \spl & \ndtw & \sr & \spl & \ndtw \\ \toprule
all data (79k) & $23.51$ & $21.94 $ & $ 40.38 $ & $ 49.96 $ & $ 47.39 $ & $ 60.80 $\\
random 0.7 & $ \bf{24.63} $ & $ \bf{23.24} $ & $ \bf{42.61} $ & $ \bf{51.63} $ & $ \bf{49.20} $ & $ \bf{62.28} $ \\ \hline
cut\_amb 0.1 & $ 22.74 $ & $ 21.01 $ & $ 40.28 $ & $ \bf{51.62} $ & $ \bf{48.82} $ & $ \bf{61.84} $ \\
cut\_amb 0.3 & $ 19.10 $ & $ 17.47 $ & $ 35.99 $ & $ \bf{50.35} $ & $ 47.31 $ & $ \bf{61.12} $ \\
cut\_amb 0.5 & $ 13.54 $ & $ 12.45 $ & $ 30.22 $ & $ 48.01 $ & $ 44.98 $ & $ 60.17 $ \\ \hline
top\_amb 0.3 & $ 22.17 $ & $ 20.30 $ & $ 39.02 $ & $ 44.29 $ & $ 42.00 $ & $ 57.50 $ \\
top\_amb 0.5 & $ 22.69 $ & $ 21.40 $ & $ \bf{41.03} $ & $ 46.93 $ & $ 44.78 $ & $ 59.59 $ \\
top\_amb 0.7 & $ \bf{24.25} $ & $ \mathbf{22.89} $ & $ \bf{41.87} $ & $ 49.50 $ & $ 47.02 $ & $ \bf{60.85} $ \\ \hline
top\_conf 0.3 & $ 21.70 $ & $ 20.17 $ & $ 39.29 $ & $ 43.80 $ & $ 41.45 $ & $ 56.89 $ \\
top\_conf 0.5 & $ \bf{24.14} $ & $ \bf{22.52} $ & $ \bf{40.83} $ & $ 47.80 $ & $ 45.11 $ & $ 59.59 $ \\
top\_conf 0.7 & $ \bf{24.70} $ & $ \bf{23.05} $ & $ \bf{41.33} $ & $ 49.44 $ & $ 46.80 $ & $ \bf{61.06} $ \\ \bottomrule
\end{tabular}
\end{table*}

\begin{table*}
\caption{Results for R2R data editing for training HAMT model from scratch and initializing from pretrained HAMT. We use overall path probability to compute confidence and variability, based on the best epoch of the model trained on full data without pretraining. ``random 0.x'' refers to training on 0.x fraction of data. ``cut\_amb 0.x'' refers to removing 0.x fraction of data with the highest variability, ``top\_amb 0.x'' refers to training on 0.x fraction of data with the highest variability. 
\textbf{Bold} represents all results, better than full data training. `-' are runs yet to complete.}
\label{tab:r2r_remove}
\tabcolsep 12pt
\centering
\begin{tabular}{lrrrrrr}
\textbf{Training Data Fraction}  & \multicolumn{3}{c}{\textbf{No pretraining}} & \multicolumn{3}{c}{\textbf{With pretraining}} \\
\cmidrule(lr){2-4}\cmidrule(lr){5-7}
 & \sr & \spl & \ndtw & \sr & \spl & \ndtw \\ \toprule
all data (14k) & $ 48.11 $ & $ 45.26 $ & $ 60.15 $ & $ 57.30 $ & $ 54.84 $ & $ 66.34 $ \\
random 0.7 & $ 47.08 $ & $ 44.35	$ & $ 58.48	 $ & $ \bf 57.68 $ & $ \bf 55.03 $ & $ \bf 66.42 $ \\ \hline
cut\_amb 0.1 & $ 46.83 $ & $ 43.90 $ & $ 58.39 $ & $ 57.00 $ & $ 54.57 $ & $ 66.23 $ \\
cut\_amb 0.3 & $ 41.51 $ & $ 39.30 $ & $ 56.44 $ & $ 56.75 $ & $ 53.98 $ & $ 65.52 $ \\
cut\_amb 0.5 & $ 38.70 $ & $ 36.31 $ & $ 53.80 $ & $ 56.15 $ & $ 53.44 $ & $ 65.32 $ \\ \hline
top\_amb 0.3 & $ 43.00 $ & $ 40.34 $ & $ 56.30 $ & $ 55.30 $ & $ 53.05 $ & $ 65.01 $ \\
top\_amb 0.5 & $ 46.45 $ & $ 43.95 $ & $ 59.63 $ & $ 56.36 $ & $ 53.77 $ & $ 65.57 $ \\
top\_amb 0.7 & $ 47.51 $ & $ 44.95 $ & $ \bf 60.83 $ & $ 57.09 $ & $ 54.43 $ & $ 66.05$ \\ 
top\_amb 0.9 & $ 47.77 $ & $ \bf 45.47 $ & $ \bf 60.45 $ & $ \bf 57.30 $ & $ \bf 55.00 $ & $ \bf 66.58 $ \\
\bottomrule
\end{tabular}
\end{table*}

\subsection{Selected Data Training Results} We report cartographs in Fig.~\ref{fig:dc_r2r},~\ref{fig:dc_rxr}, for R2R and RxR, for the best validation unseen performance. For RxR, the dense hard-to-learn regions on the maps show that the multilingual \vln task is very hard to learn and generalize to new environments and instructions, as it requires strong language grounding in perception and actions. We also report a data map for RxR without pretraining in Fig.~\ref{fig:dc_rxr_wo_pt}. We find that the task is even harder compared to with pretraining case. Table~\ref{tab:rxr_remove} and Table~\ref{tab:r2r_remove} report results when training with different data editing scenarios on RxR and R2R, based on the data map in Fig.~\ref{fig:dc_rxr} and Fig.~\ref{fig:dc_r2r},  correspondingly. Training on the same data splits when initialized from the pretrained \hamt model leads to different result from in~\citet{swayamdipta2020dataset}.

\paragraph{Pretraining is the dominant stage for R2R}
From the pretraining results in Table~\ref{tab:r2r_remove}, we observe it is hard to either increase or decrease the performance by selected data training, indicating the \vln pretraining is the dominant stage which influences the performance on R2R.

\paragraph{Pretrained model on RxR is almost saturated}
Cutting out highest ambiguity data leads to better model performance, in comparison to full data training. Thus, the low confidence, low variability regions become the good data region for with pretraining case. 
However, being more centralized to the ``good data region'' by selecting the top ambiguous data, can only reach on-par or lower performance than the full data training instead of further improving the performance.
This shows the pretrained model is almost saturated, which can be hard to improve but easy to contaminate.

\paragraph{Data cleaning is less effective in \vln finetuning}

The initial results of pretraining indicate removing ambiguous points adds the most to the model performance, which is in direct contradiction to~\cite{swayamdipta2020dataset}. 
We conducted some experiments where we trained the model from scratch without pretraining to investigate this.
Fig.~\ref{fig:dc_rxr_wo_pt}, as we suspected that the pretraining on synthetically augmentation data maybe affecting the model behavior in a way that is making it difficult to study the true response to different data regions. 

We have similar findings in both Table~\ref{tab:r2r_remove} and~\ref{tab:rxr_remove}.  
We find that without pretraining, the high confidence and high variability data adds the most to the model performance. 
The model consistently improves on all metric as we increase the amount of high confidence or high variability data. 
Consequently, the model consistently and rapidly degrades as we cut out these regions of data. 
Training with high confidence, high variability data even outperforms the training run with full dataset. 

The random data selection leads to the beast performance overall, in the both cases. 
To verify this result, we repeated the experiment multiple time with different sets of random data selection. 
Excluding the random and all data runs, \textit{increasing the amount of data always improves performance}.

Therefore, our finding from cartography analysis are contradictory to that in~\cite{swayamdipta2020dataset}, which shows that extracting out good data improves model performance, and concludes that data quality matters more than data quantity. This finding motivates and sets the basis of our above research on data quality \vs quantity.

\end{document}